\documentclass{article}
\usepackage{spconf,amsmath,graphicx,multirow}

\title{Window-Based Descriptors for Arabic Handwritten Alphabet Recognition: A Comparative Study on A Novel Dataset}
\name{Marwan Torki\sthanks{Contact author's email: mtorki@alexu.edu.eg}, Mohamed E. Hussein\sthanks{Mohamed E. Hussein is currently an Assistant Professor at Egypt-Japan University of Science and Technology, New Borg El-Arab City, Alexandria, Egypt; on leave from his position at Alexandria University, where most of the work associated with this project was performed.}, Ahmed Elsallamy, Mahmoud Fayyaz, Shehab Yaser}
\address{Computer and Systems Engineering Department, \\
Faculty of Engineering, Alexandria University}
\begin{document}
%\ninept
%
\maketitle
\begin{abstract}
This  paper  presents  a  comparative  study  for window-based descriptors on the application of Arabic handwritten alphabet recognition. We show a detailed experimental evaluation of different descriptors with several classifiers. The objective of the paper is to evaluate different window-based descriptors on the problem of Arabic letter recognition. Our experiments clearly show that they perform very well. Moreover, we introduce a novel spatial pyramid partitioning scheme that enhances the recognition accuracy for most descriptors. In addition, we introduce a novel dataset for Arabic handwritten isolated alphabet letters, which can serve as a benchmark for future research.

\end{abstract}
\section{Introduction}
\label{sec:intro}
The interest in Arabic handwritten recognition has increased during the last few years~\cite{parvez_offline_2013}. %Papers focused on Arabic are doubled since the ICFHR 2010 to ICFHR 2012. 
The successful recognition of Arabic handwritten alphabet is an important milestone towards the goal of a successful Optical Handwriting Recognition (OHR) system. 

Window-based descriptors have shown a great success in the recent computer vision research on object recognition and detection ~\cite{szeliski2010computer}. Many of the window-based descriptors depend on the gradients in the window of interest, or depend on the texture. The intuition behind using window-based descriptors on the problem of character recognition is that the shapes of the letters can be represented using lines, circles, semi-circular curves, and angles. The composition of character shapes from such primitive shapes suggests that gradient and texture features could be used to represent character shapes for the purpose character recognition. We show in this paper that existing window-based descriptors (which are based on gradient and texture features) can be effectively used in the context of letter recognition for Arabic handwriting.

The contribution of this paper is multifold. \textit{First}, the paper introduces a compact 9K novel dataset of 28 classes that represent the isolated Arabic handwritten alphabet. The dataset's ground truth annotation includes writers' genders. \textit{Second} the paper presents a comparative evaluation of common window based descriptors in computer vision literature, namely, HOG, SIFT, SURF, LBP,  and GIST. \textit{Third}, the paper presents a comparative evaluation of four common classifiers on the chosen descriptors, namely, Logistic Regression, Linear SVM , Non-linear SVM, and Artificial Neural Networks. 

The remainder of this paper is organized as follows: Section \ref{sec:RW} gives an overview of gradient based window descriptors as well as texture based window descriptors. Afterwords, our novel dataset for isolated Arabic alphabet dataset will be introduced in Section \ref{sec:data9k}. Experimental evaluation will be presented in Section \ref{sec:expr}. The conclusion is given in Section \ref{sec:conc}.

\section{Background}
\label{sec:RW}
\begin{figure}[t]
\begin{minipage}[b]{1\linewidth}
  \centering
  \centerline{\includegraphics[width=.8\linewidth,height=.750cm]{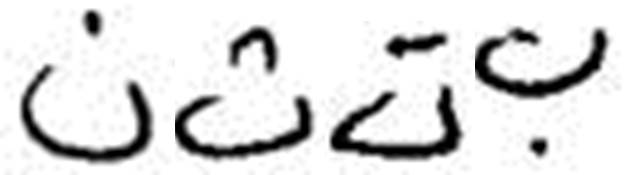}}
%  \vspace{1.5cm}
  \centerline{(a)Sample confusing classes}\medskip
\end{minipage}

\begin{minipage}[b]{1\linewidth}
  \centering
  \centerline{\includegraphics[width=.9\linewidth,height=1.0cm]{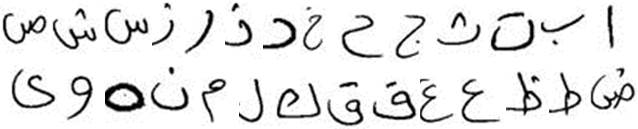}}
%  \vspace{1.5cm}
  \centerline{(b) Arabic Alphabet}\medskip
\end{minipage}
\caption{Confusing classes share layout and shape(Top). Full Alphabet(Bottom) shows lots of confusing classes}
\label{fig:Alphabet}
\end{figure}

The Arabic alphabet is widely used in many countries including all Arabic speaking countries. Other languages that use the same alphabet are Persian, Pashto, and Urdu \footnote{$http://en.wikipedia.org/wiki/Arabic\_script$}. The wide use of these languages adds importance to the Arabic alphabet recognition problem. The difficulty of Arabic alphabet recognition is due to the fact that many characters have similar shapes, outlines, and dots. Figure ~\ref{fig:Alphabet}.a shows four different classes and they almost have the same shape and outline; and the only differences are in the dots' locations and numbers. The number of classes in the Arabic alphabet is 28. As of figure~\ref{fig:Alphabet}, it can be seen that the confusion between certain classes is very common such that almost half of the letters are having confusing class(es) with only the number and/or location of dots changed. Much research has been conducted on feature extraction methods like[]. None of these methods used the common window-based descriptors in the object/scene recognition literature, such as HOG~\cite{Dalal05}, SIFT~\cite{Lowe04}, LBP~\cite{LBP02}, GIST~\cite{Torralba03} and SURF~\cite{Bay08}.\\
\subsection{Window-Based Descriptors:} An example of a successful window-based descriptor is the Histogram of Oriented Gradients (HOG)~\cite{Dalal05}. Scale Invariant Feature Transform (SIFT)~\cite{Lowe04}, and Speeded-Up Robust Features (SURF)~\cite{Bay08} are considered patch-based descriptors but since we are dealing with the whole image as a single patch we can consider them as window-based descriptors as well. 

The HOG descriptor bins the oriented gradients within overlapping cells. The SIFT descriptor consists of a histogram of oriented image gradients captured within grid cells within a local region. The SURF descriptor is an efficient alternative to SIFT that uses simple 2D box filters to approximate derivatives and it uses summary statistics instead of using histogram counts.

Another set of descriptors describe the texture pattern in a window instead of gradient orientations. One of these descriptors is the GIST descriptor~\cite{Torralba03}. GIST applies a set of filters on a 4x4 cell grid. Within each cell, it records orientation histograms computed from Gabor filter responses. Another texture-based descriptor is the Local Binary Pattern descriptor ~\cite{LBP02} .\\
\subsection{Spatial Pyramid:} \label{sec:spatialp}
 Spatial pyramid of descriptors showed improvement on recognition problems over the original descriptors~\cite{Laze,Bosch07}. Inspired by the spatial pyramid, a temporal pyramid was introduced in activity recognition problems~\cite{ours1} where the temporal dependencies are described by the pyramid. Moreover, allowing for overlapping between temporal windows further improved the recognition accuracy ~\cite{ours1}. 

\begin{figure}[htb]
\begin{center}
\includegraphics[width=1\linewidth]{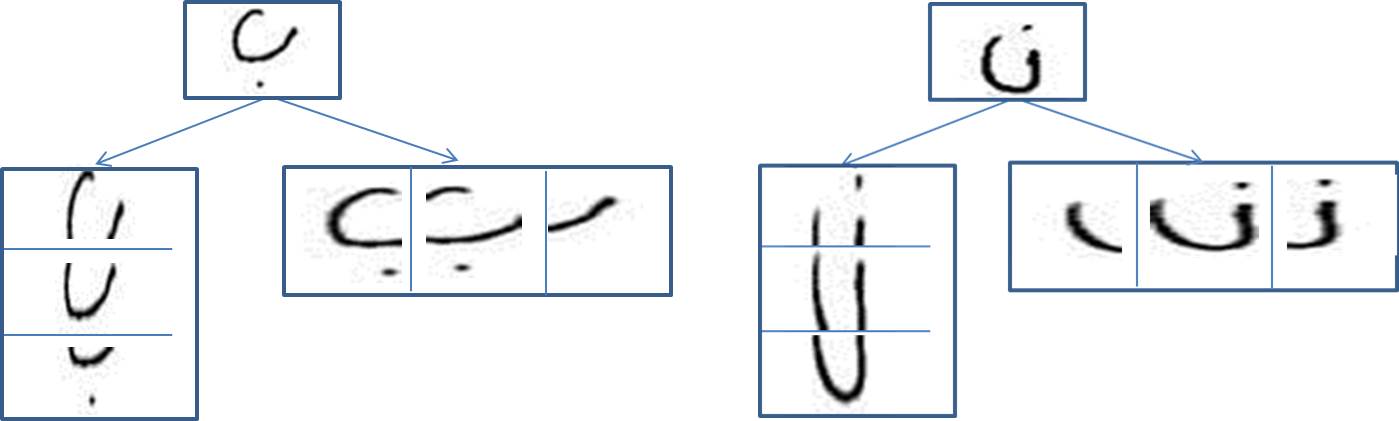}
\caption{Role of overlapped spatial partitioning on letter recognition in discriminating confusing classes}
\label{fig:batathanoonrev}
\end{center}
\end{figure}

Success of overlapping temporal pyramid in activity recognition and the structure of the Arabic alphabet lead us to introduce the spatially overlapping version of the window based descriptors. The effect of adding the overlapping spatial partitioning is illustrated in Figure~\ref{fig:batathanoonrev}, the different parts of the resulting descriptor will be much better discriminating the confusing classes. We propose to use three overlapping halves of the image of the character in each direction in addition to the original image. Thus every image is represented by a concatenation of 7 descriptors one for the original image, three overlapping halves in the vertical direction and three overlapping halves in the horizontal direction. We call the new descriptors HOG7, GIST7, LBP7, SIFT7 and SURF7. Experimental evaluation in sec.~\ref{sec:expr} shows that there can be a significant improvement in the recognition accuracy when using the descriptors with these spatial pyramids over using the original descriptors.

\begin{figure*}[t]
\begin{minipage}[a]{.48\linewidth}
  \centering
  \centerline{\includegraphics[height=10.0cm,width=8.0cm]{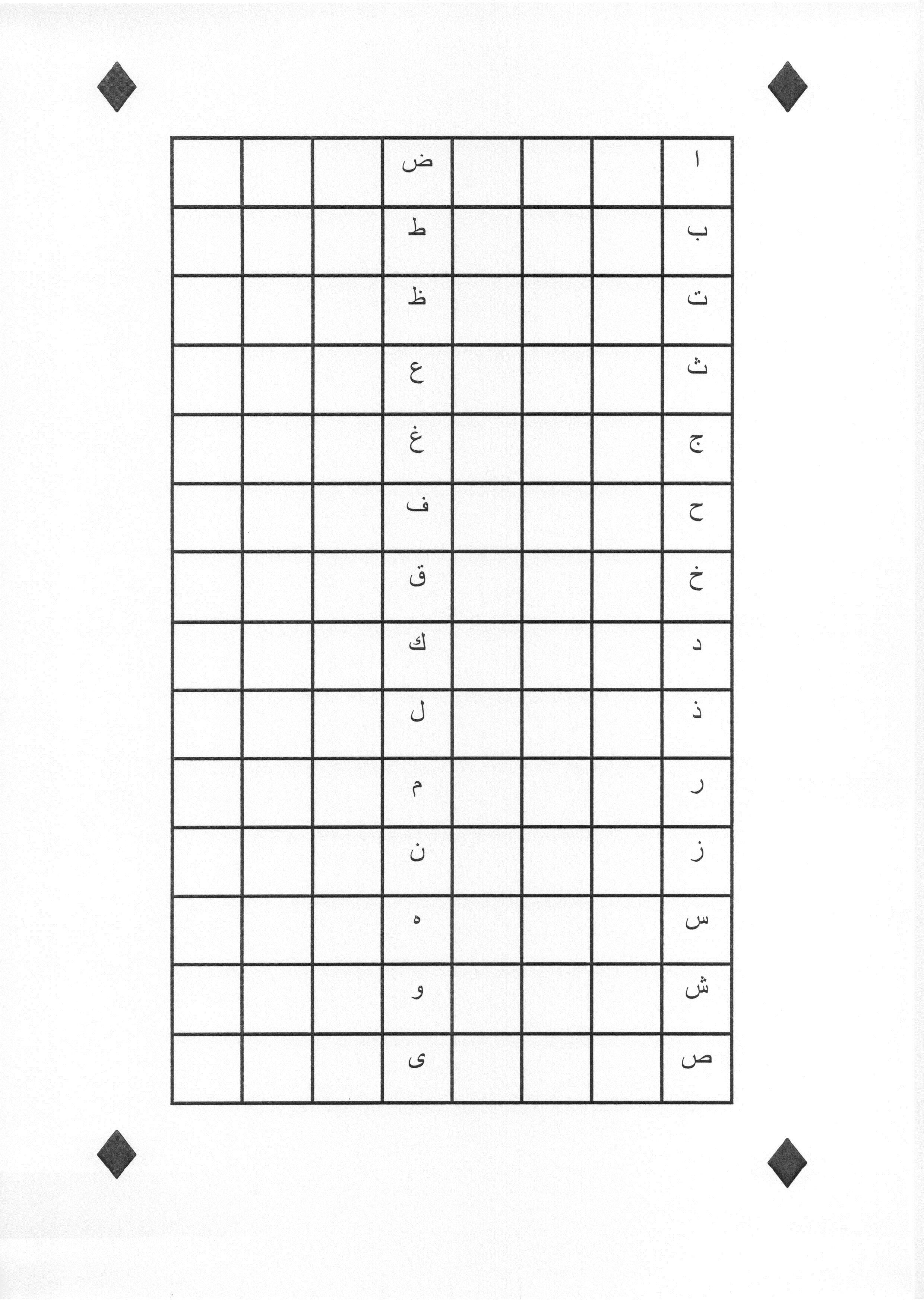}}
%  \vspace{1.5cm}
  \centerline{(a) Empty form}\medskip
\end{minipage}
\hfill
\begin{minipage}[a]{0.48\linewidth}
  \centering
  \centerline{\includegraphics[height=10.0cm,width=8.0cm]{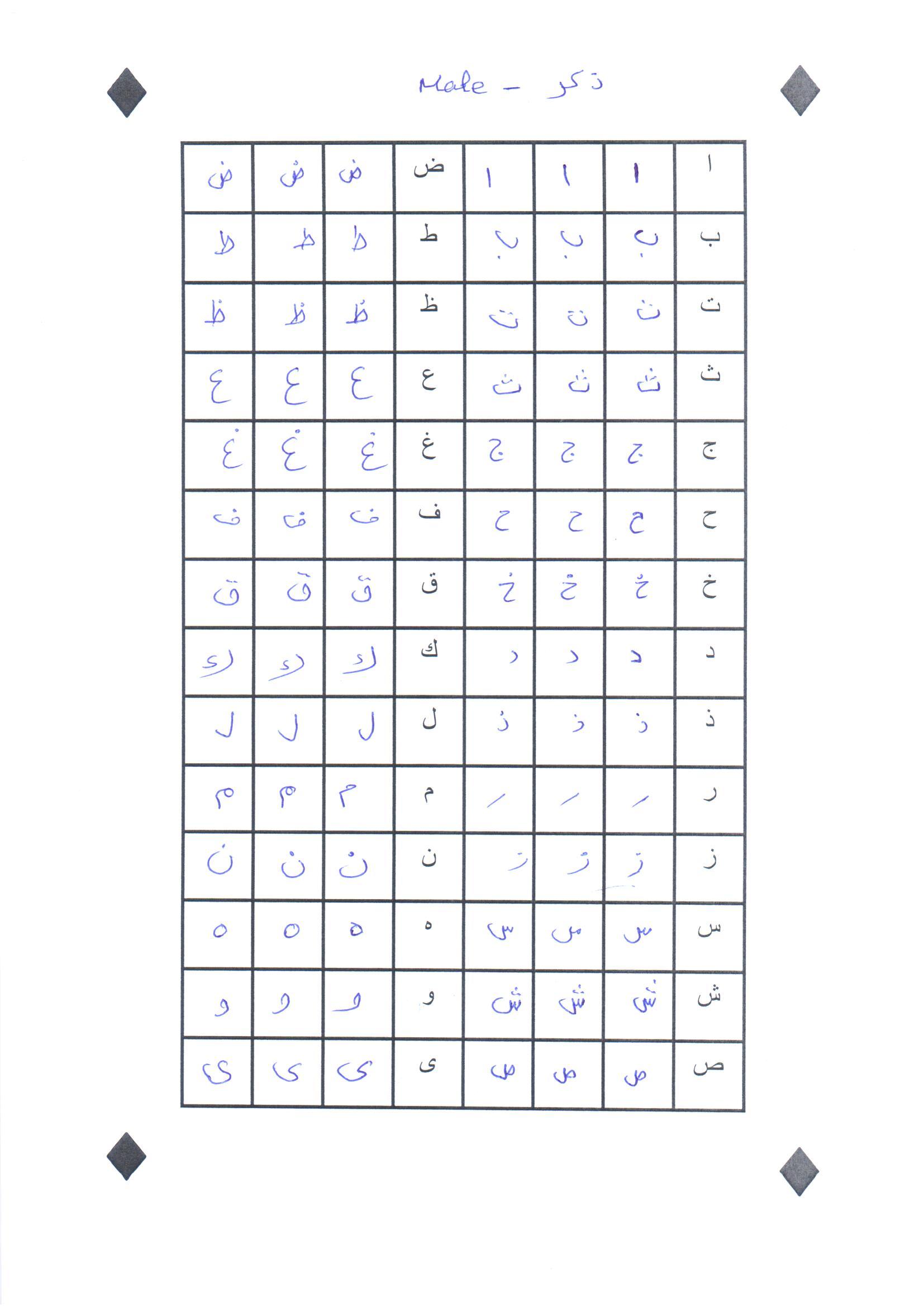}}
%  \vspace{1.5cm}
  \centerline{(b) Filled form}\medskip
\end{minipage}
\caption{Form used in the collection}
\label{fig:form}
\end{figure*}
\section{AIA9k dataset}
\label{sec:data9k}
AlexU Isolated Alphabet (AIA9K) dataset \footnote{Dataset is available for download by contacting first author mtorki@alexu.edu.eg} was collected from 107 volunteer writers, between 18 to 25 years old, and who are B.Sc. or M. Sc. students in the Faculty of Engineering at Alexandria University. The 107 writers included 62 females and 45 males. Each writer wrote all the Arabic letters 3 times. The total number of collected characters is 8,988 letters. Figure~\ref{fig:form} shows the form that has been used to collect the data.

To obtain the images of the isolated characters from the scanned form image, some form pre-processing steps were performed. These steps included Harris corner detection and finding the geometric transformation to handle skewness in the scanned forms. Afterward a border removal and cell cropping steps were applied.

Each Writer was asked to either leave his/her name (from which the gender was inferred) or write Male/female on the form. Forms without such information (only two) were excluded.

\begin{figure}[htbp]
	\begin{minipage}[b]{.19\linewidth}
  \centering
  \centerline{\includegraphics[width=1.0\linewidth,height=.4\linewidth]{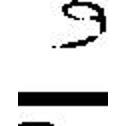}}
%  \vspace{1.5cm}
  \centerline{(a)}\medskip
  \end{minipage}
  \hfill
  \begin{minipage}[b]{0.19\linewidth}
  \centering
  \centerline{\includegraphics[width=1.0\linewidth,height=.4\linewidth]{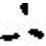}}
%  \vspace{1.5cm}
  \centerline{(b)}\medskip
  \end{minipage}
  \begin{minipage}[b]{.19\linewidth}
  \centering
  \centerline{\includegraphics[width=1.0\linewidth,height=.4\linewidth]{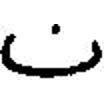}}
%  \vspace{1.5cm}
  \centerline{(c)}\medskip
  \end{minipage}
  \hfill
  \begin{minipage}[b]{0.19\linewidth}
  \centering
  \centerline{\includegraphics[width=1.0\linewidth,height=.4\linewidth]{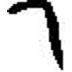}}
%  \vspace{1.5cm}
  \centerline{(d)}\medskip
	
  \end{minipage}
	\begin{minipage}[b]{0.19\linewidth}
  \centering
  \centerline{\includegraphics[width=1.0\linewidth,height=.4\linewidth]{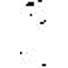}}
%  \vspace{1.5cm}
  \centerline{(e)}\medskip
	
  \end{minipage}	
	\caption{Three types of errors: Cropping(a,b), Writer mistakes(c,d) and Unclear letters(e).}
	\label{fig:errors}
\end{figure}

A verification process for the ground truth was executed; and three types of errors were identified, as follows.\\
 \textbf{(1)Cropping Error}: Some letters are wrongly cropped, producing more connected components in the image, as in ~\ref{fig:errors}.a, or only a part of the letter, as in Figure~\ref{fig:errors}.b \textbf{(2)Writer Mistake}: This occurs when a writer writes a letter in the wrong place, producing a wrong label. Missing points is a very common case, as in Figure~\ref{fig:errors}.c. Some writing is not a letter at all, as in Figure~\ref{fig:errors}.d. \textbf{(3)Unclear Letters}: This usually happens due to scanning errors, or writing errors as in Figure~\ref{fig:errors}.e. If the unclear part covered most of the main features of the letter, the letter is removed. The verification process resulted in downsizing the data to 8737 valid samples.

The training/ testing split was $70\%$ for training, $15\%$ for validation, and  $15\%$ for testing. The validation set is used to find the best parameters for each classifier. Later the training set is combined with the validation set to perform the final classification on the testing set. Table~\ref{tab:split} shows the exact split, which takes writers' genders into account.
\begin{table}[htbp]
  \footnotesize
	\centering
  \caption{Data Splits}
    \begin{tabular}{ccccc}
    \hline
    \textbf{} & \textbf{Total} & \textbf{70\% Train} & \textbf{15\% Validation} & \textbf{15\% Test} \\\hline
    \textbf{Female} & 5089  & 3562  & 763   & 764 \\\hline
    \textbf{Male} & 3648  & 2553  & 547   & 548 \\\hline
    \textbf{Total} & 8737  & 6115  & 1310  & 1312 \\\hline
    \end{tabular}%
  \label{tab:split}%
\end{table}%
\section{Experiments}
\label{sec:expr}
In this section we show our experiments on AIA9K dataset. Experiments are designed   to recognize the letters with different descriptors and classifiers. 
\subsection{Classifier Tuning on Validation Set}
We used the validation set to adjust the parameters of different classifiers: the regularization parameter ($\lambda$) of the Logistic Regression (LR) classifier, the regularization parameter ($\lambda$) of Artificial Neural Network (ANN) (only one hidden layer of 25 neurons was used in all experiments), the $C$ parameter for the Support Vector Machine (SVM) with linear kernel, and finally, the $\gamma$ and $C$ parameters for the SVM with RBF kernel.

\subsection{Descriptors}
We used five different descriptors: three gradient-based descriptors (HOG, SIFT, and SURF), and two texture-based descriptors (GIST and LBP). To benefit from the spatial layout of the characters, we used a modified spatial partitioning that allows overlapping, presented in Section~\ref{sec:spatialp}, and we appended the name of each descriptor by number 7 to differentiate it from the original descriptor.

\subsection{Results: Character Recognition}
\begin{figure}[htb]
\begin{center}
\includegraphics[width=1\linewidth]{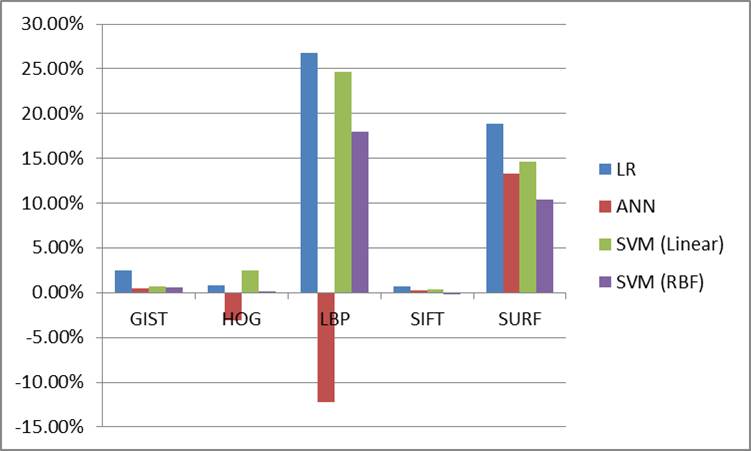}
\caption{Improvements using the overlapped spatial partitioning on letter recognition}
\label{fig:overSpaPyr}
\end{center}
\end{figure}

\begin{figure}[htb]
\begin{center}
\includegraphics[width=.8\linewidth]{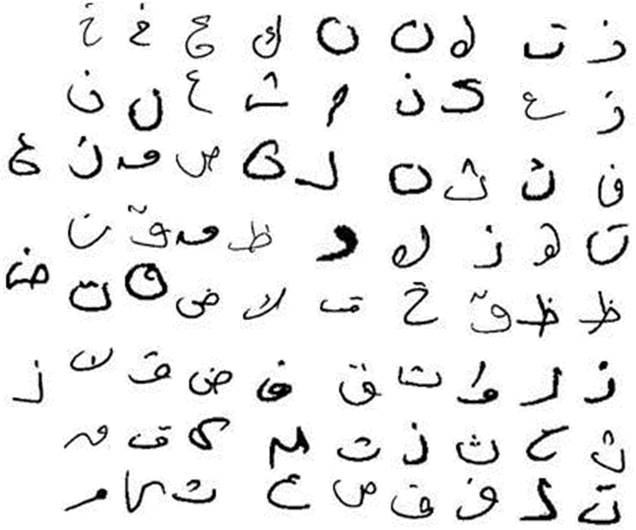}
\caption{75 miss-classified letters using SIFT descriptor and SVM(RBF)}
\label{fig:missed}
\end{center}
\end{figure}

Table~\ref{tab:mainres} shows the recognition rates using different classifiers and descriptors. It can be seen that SIFT is doing the best with and without spatial overlapping partitioning. As in Figure~\ref{fig:overSpaPyr}, it can be seen the the spatial overlapping partitioning, in general, improves the results. An interesting result is that the LBP feature is performing the worst and also it is interesting that spatial overlapping partitioning increased LBP's recognition rate the most. Figure~\ref{fig:missed} shows the 75 miss-classified letters by our best configuration, using the SIFT descriptor with SVM (RBF). 
% Table generated by Excel2LaTeX from sheet 'Sheet1'
\begin{table}[htbp]
  \footnotesize
	\centering
  \caption{Accuracy using Train+Validation sets for training and Test set for testing: Character recognition}
    \begin{tabular}{lcccc}
    \hline
    \textbf{} & \textbf{LR} & \textbf{ANN} & \textbf{SVM (Linear)} & \textbf{SVM (RBF)} \\
    \hline
    \textbf{GIST} & 88.26\% & 91.01\% & 90.63\% & 92.61\% \\\hline
    \textbf{HOG} & 86.51\% & 86.66\% & 87.25\% & 90.46\% \\\hline
    \textbf{LBP} & 52.97\% & 57.09\% & 55.03\% & 57.32\% \\\hline
    \textbf{SIFT} & 92.07\% & 92.99\% & 92.45\% & \textbf{94.28\%} \\\hline
    \textbf{SURF} & 66.39\% & 72.64\% & 70.05\% & 77.21\% \\\hline\hline
    \textbf{GIST7} & 90.70\% & 91.54\% & 91.31\% & 93.22 \\\hline
    \textbf{HOG7} & 87.27\% & 83.61\% & 89.71\% & 90.47\% \\\hline
		\textbf{LBP7} & 79.73\% & 44.89\% & 79.65\% & 75.30\% \\\hline
    \textbf{SIFT7} & 92.76\% & 93.29\% & 92.84\% & \textbf{94.13\%} \\\hline
    \textbf{SURF7} & 85.21\% & 85.90\% & 84.68\% & 87.58\% \\\hline
    \end{tabular}%
  \label{tab:mainres}%
\end{table}%

\section{Conclusion}
\label{sec:conc}
In this paper, we introduced a novel dataset (AIA9K) for isolated Arabic alphabet characters. We also performed a quantitative evaluation for window-based descriptors on the novel dataset. The chosen descriptors showed competitive recognition rates except for the LBP and SURF. SIFT produced the best accuracy (only 75 samples out of 1312 were miss-classified). A remarkable improvement on different descriptors was obtained using overlapped spatial partitioning of the character image. 

% use section* for acknowledgement
\section*{Acknowledgment}
The authors would like to thank the Center of Excellence for Smart Critical Infrastructure (SmartCI) at Virginia Tech - Middle East and North Africa (VT-MENA) for sponsoring this work. The authors would also like to thank May Shoushan, Eiman El Gamel, Asmaa Shoala, Noha Ali, and Ayat Abdel-Fattah for their effort in data collection.

\bibliographystyle{IEEEbib}
\footnotesize
%\bibliography{refs}

\end{document}